\documentclass{article}
\usepackage{spconf,amsmath,graphicx}

\usepackage{enumitem}
\setlist{nosep, leftmargin=14pt}
\usepackage{hyperref}
\usepackage{mwe} 

\title{Teeth and root canals segmentation using ZXYformer with uncertainty guidance and weight transfer}
\name{Shangxuan Li$^{1,\dagger}$,~ Yu Du$^{2,3,4,\dagger}$,~ Li Ye$^{2,3,4}$,~ Chichi Li$^{5}$,~ Yanshu Fang$^{6}$,~ Cheng Wang$^{5,\star}$,~ Wu Zhou$^{1,\star}$ \thanks{$\dagger$ Equal contribution $\star$ Corresponding author: Cheng Wang, cheng.wang@hanglok-tech.cn; Wu Zhou, zhouwu@gzucm.edu.cn. This research is supported by the school-enterprise cooperation project (No.6401-222-127-001).}}

\address{$^{1}$ School of Medical Information Engineering, Guangzhou University of Chinese Medicine, China \\
$^{2}$ Department of Operative Dentistry and Endodontics, Sun Yat-sen University, Guangzhou, China\\
$^{3}$ Affiliated Stomatological Hospital, Guangzhou, China\\
$^{4}$ Guangdong Provincial Key Laboratory of Stomatology, Guangzhou, China\\
$^{5}$ Hanglok-Tech Co.,Ltd., Zhuhai, China\\
$^{6}$ First Clinical Medical College, Guangzhou University of Chinese Medicine, Guangzhou, China
}
\begin{document}
%
\maketitle
\begin{abstract}
This study attempts to segment teeth and root-canals simultaneously from CBCT images, but there are very challenging problems in this process. First, the clinical CBCT image data is very large (e.g., $672 \times 688 \times 688$), and the use of down-sampling operation will lose useful information about teeth and root canals. Second, teeth and root canals are very different in morphology, and it is difficult for a simple network to identify them precisely. In addition, there are weak edges at the tooth, between tooth and root canal, which makes it very difficult to segment such weak edges. To this end, we propose a coarse-to-fine segmentation method based on inverse feature fusion transformer and uncertainty estimation to address above challenging problems. First, we use the down-scaled volume data (e.g., $128 \times 128 \times 128$) to conduct coarse segmentation and map it to the original volume to obtain the area of teeth and root canals. Then, we design a transformer with reverse feature fusion, which can bring better segmentation effect of different morphological objects by transferring deeper features to shallow features. Finally, we design an auxiliary branch to calculate and refine the difficult areas in order to improve the weak edge segmentation performance of teeth and root canals. Through the combined tooth and root canal segmentation experiment of 157 clinical high-resolution CBCT data, it is verified that the proposed method is superior to the existing tooth or root canal segmentation methods.
\end{abstract}
\section{Introduction}
\label{sec1}

Root canal therapy is an important procedure to treat pulp and periapical diseases. Accurate identification of root canal status of affected teeth based on CBCT is of great significance to evaluate the difficulty of treatment and improve the treatment effect \cite{0dd57b9b00f14995a680f4ccd419106f}. Therefore, the simultaneous segmentation of teeth and root canals in CBCT images and the simultaneous visualization of affected teeth and root canals have important clinical value for the treatment of oral diseases.

At present, deep learning has made great progress in tooth and root canal segmentation \cite{8954147,10.1007/978-3-030-78191-0_12,article,9098542,9629727}. However, it still faces great challenges in clinical practice. First, the clinical CBCT image data is very large (e.g., $672 \times 688 \times 688$), while the size of teeth and root canals in CBCT images is small. The current deep network is difficult to directly process such large volume data, and down-sampling processing will lose useful information of teeth and root canals. The current research generally adopts to manually cut out the tooth or root canal region from the original volume \cite{8954147,10.1007/978-3-030-78191-0_12,9098542,9629727}, and then put it into the deep network for segmentation, which brings a very large manual burden to the clinical. Second, teeth and root canals are very different in morphology, and it is difficult to obtain good segmentation results using fixed segmentation network parameters. In the current research, teeth and root canals segmentation are conducted independently \cite{10.1007/978-3-030-78191-0_12,9629727}, and there is no joint segmentation of teeth and root canals. In addition, both teeth and root canals have weak edges. The current research generally uses the edge map for constraint enhancement \cite{8954147,10.1007/978-3-030-78191-0_12,article,9629727}, and does not consider the segmentation results for refinement and uncertainty analysis, which greatly reduces the segmentation performance of weak edges.

To solve the above problems, we propose a coarse-to-fine segmentation method based on inverse feature fusion transformer and uncertainty estimation. First, the area of teeth and root canals on the original image is obtained by coarse segmentation of the down-sampled volume data, which solves the problem that high-resolution CBCT cannot be directly processed in the network. Meanwhile, the coarse segmentation network parameters are migrated to the fine segmentation network to improve the performance of the fine segmentation network. Then, we design a transformer with reverse feature fusion to transfer deeper features to shallow features, so as to improve the details segmentation effect of different morphological objects. Finally, we calculate the uncertainty segmentation region, and further improve the weak edge segmentation performance of teeth and root canals. 157 clinical high-resolution CBCT data are used to assess the performance of the proposed method.

\section{Method}
\label{sec2}

\subsection{The framework of the approach}
\begin{figure*}[htb]
\centerline{\includegraphics[width=1.6\columnwidth]{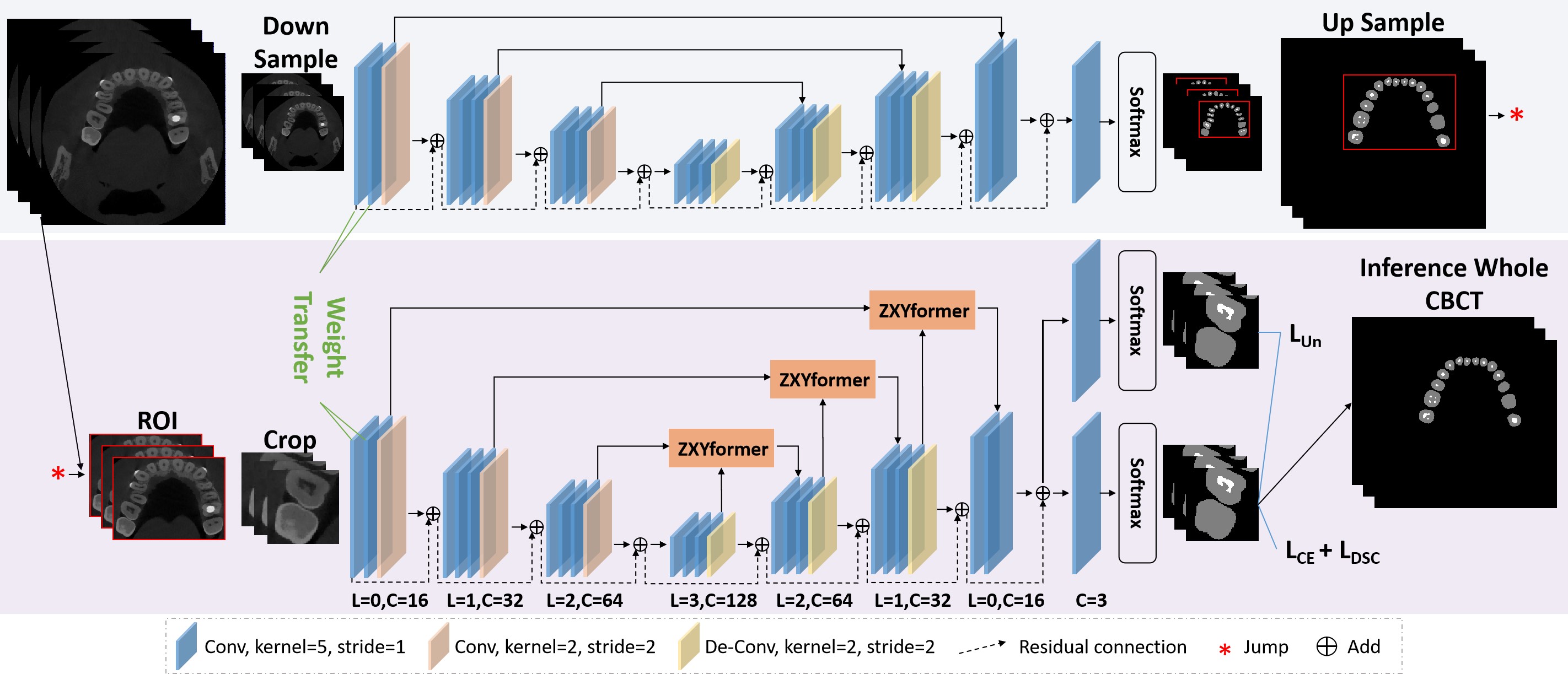}}
\caption{Network structure of tooth and root canal segmentation.}
\label{fig1}
\end{figure*}

The network structure of the proposed method is shown in Fig.\ref{fig1}, which is mainly a coarse-to-fine segmentation network, including 1)the design of weight transfer from coarse segmentation network to fine segmentation network; 2) ZXYformer: Z process is responsible for expanding channel dimensions and upsampling high-level features, X process is responsible for morphological information learning and reverse feature fusion, and Y process includes feedforward neural network (FFN) and channel dimension restoration; 3) Calculating the auxiliary branch of the segmentation edge uncertainty. First, the region of tooth and root canal on the original image is obtained by coarse segmentation of the down-sampled volume data, and the coarse segmentation network parameters are transferred to the fine segmentation network. Then, ZXYformer is designed to capture global features and multi-channel features, followed by applying the upper sampling layer and convolution layer to gradually produce high-resolution segmentation results. Finally, an auxiliary branch is designed in order to improve the weak edge segmentation performance of teeth and root canals. We will introduce the modules in detail in the following subsections.

\subsection{Weight transfer}
In order to obtain tooth and root canal regions to be segmented from high-resolution CBCT images, we design a coarse-to-fine segmentation network. First, we reshape CBCT images to $128 \times 128 \times 128$ to segment teeth and root canals. Then, we migrate the shallow coarse segmentation network to fine segmentation network to maintain the macro information in coarse network. Unlike \cite{article}, which extracts patches directly from the original image at random, the proposed method enhances the fine segmentation network while using the weight migration method without losing macro information.

\subsection{Encoder-Decoder}
We ramdonly extract size of $128 \times 128 \times 128$ patch in the region of coarse segmentation results mapping to the original volume, and use 3D convolution layer to extract local information of teeth and root canalse, gradually encoding $F$ to rich local context and high-level feature. We input feature $F$ into ZXYformer encoder to further learn the long-distance correlation of global receptive field. In the decoding phase, we perform convolutional upsampling and cascading operations. We use skip connections to fuse the characteristics of the encoder and decoder. In the process of skip connections, we use the proposed ZXYformer to achieve a more detailed segmentation mask.

\subsection{ZXYformer feature embedding}
\begin{figure}[htb]
\centerline{\includegraphics[width=0.8\columnwidth]{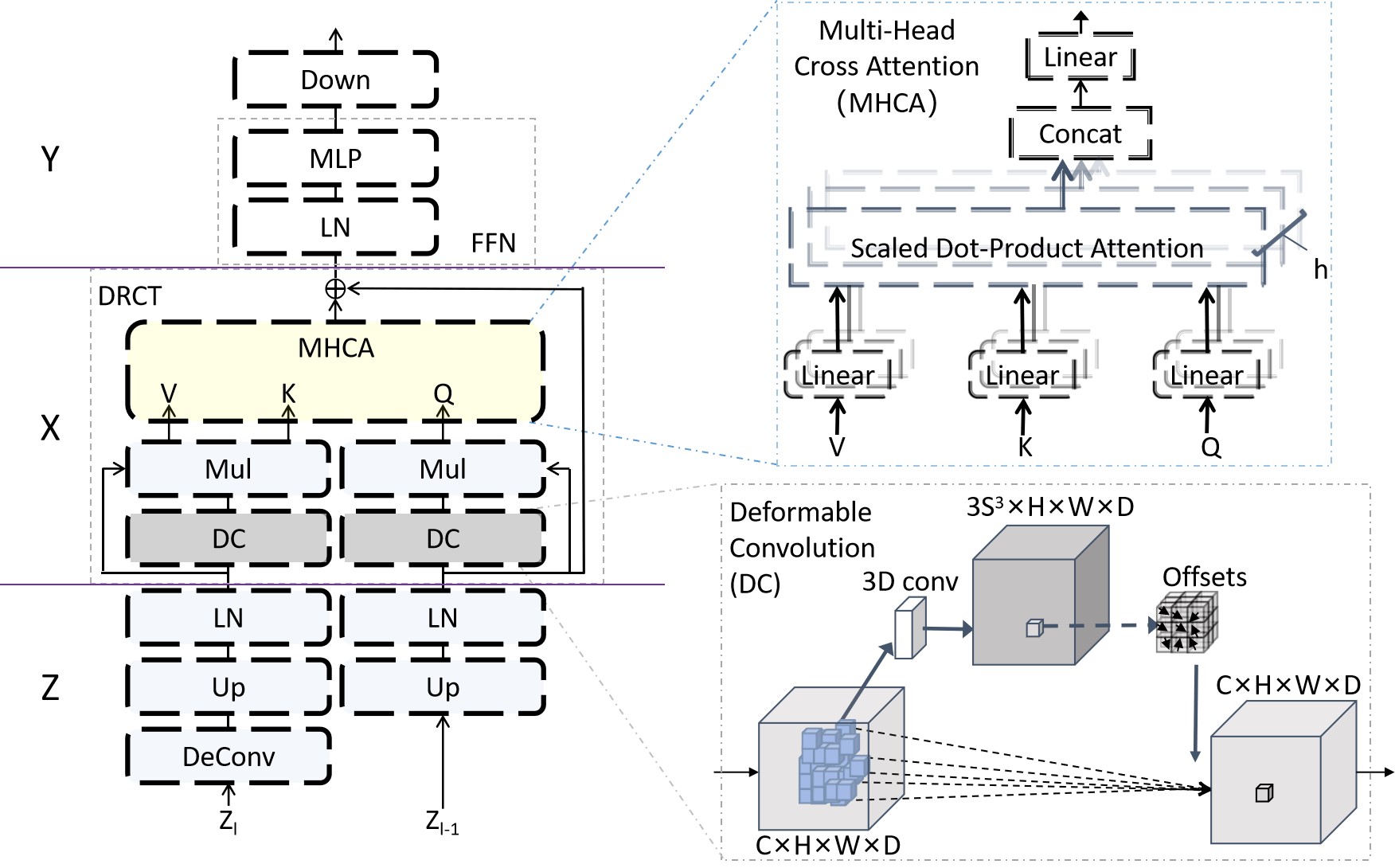}}
\caption{Structure of ZXYformer. DeConv is deconvolution, Up and Down use $1\times1\times1$ convolution to separately increase and decrease the dimension of the channel, LN represents layer normalization, Mul represents matrix dot multiplication, MLP is multilayer perceptron.}
\label{fig2}
\end{figure}
As the size of the area felt by different level features is inconsistent, transferring the wider field of vision information to the local field of vision can bring better segmentation details \cite{9759447}, especially for slender structures \cite{XIA2022102581}. Therefore, we design a transformer with reverse feature fusion to transfer deeper features morphological information to shallow features. As shown in Fig.\ref{fig2}, the proposed ZXYformer consists of three processes, in which the Z process upsamples features of high levels, and expands and regularizes the channel dimensions of two features. The X process is a Deformable reverse cross transformer (DRCT), and the Y process includes a FFN and channel dimension restoration. 

Specifically, we introduc deformable convolution (DC)\cite{8237351} to capture the shape perception features of irregular shaped lesions, which has been proved to be a shape perception module. In addition, morphological information from different levels will use a multi-head cross-attention to guide shallow detail learning with deeper information, in order to constrain the edge voxels similar to vascular structures and improve segmentation performance. Then, the Y process uses the conventional FFN and restores the channel dimension. Finally, through the expansion and compression of the channel dimension of the Z and Y processes, the width of the transformer can be expanded to ensure that each voxel can be fully represented.

\subsection{Uncertainty estimation}
As the weak contrast between the root edge of the tooth and the alveolar bone, the calcification of the root canal, tooth decay and other conditions, there are many uncertain regions in the simultaneous segmentation of tooth and root canal in CBCT. We design an auxiliary branch, and use the predicted  Kullback-Leibler divergence as the variance to calculate the uncertainty \cite{10.1007/s11263-020-01395-y}. If two classifiers provide different class predictions, the approximate variance will get a larger value. It reflects the uncertainty of the model on prediction:
\begin{equation}
L_{Un}=exp(p_{aux}~log(\frac{p_{main}}{p_{aux}}))
\end{equation}
where $p_{main}$ and $p_{aux}$ represent the prediction probabilities of the main branch and the auxiliary branch respectively.

On the main branch, we apply the conventional cross entropy loss ($L_{CE}$) and dice loss ($L_{Dice}$) to the prediction to optimize the model. The total loss function of the proposed model to be optimized can be defined as: $L_{Total}=L_{CE}+L_{Dice}+L_{Un}$. The training of difficult areas is corrected without introducing additional parameters or modules by minimizing the uncertainty of segmentation prediction.

\subsection{Implementation}
The initialization setting of the learning rate is 1e-3, with 50000 iterations. The Adam algorithm is used to minimize the objective function. Two RTX3090 GPUs are used, each with 24G memory. The attenuation setting of the learning rate is 0.99 for every 500 iterations. All parameters, including weights and deviations, are initialized using a truncated normal distribution with a standard deviation of 0.1. We use connected component analysis to extract the maximum area of predicted voxels and remove some small false-positive voxels. The basic implementation code of this work will be shared in GitHub, \url{https://github.com/LSX0802/ZXY}.

\section{Experimental results and discussion}
\label{sec3}

\subsection{Clinical data, pre-processing and evalutaion metrics}
This study was performed in line with the principles of the Declaration of Helsinki. Approval was granted by the Ethics Committee of my institution. In the experiment, 157 cases of CBCT data from the medical imaging department of the local hospital were used. The dental symptoms of 157 cases were recognized as oral CBCT indications by professional stomatologists. The data image resolution of the CBCT equipment used is $0.2\sim1.0 mm$. The ground truth label was marked by a physician and reviewed by two dental and endodontics specialists with 10 years of clinical experience. First, considering the balance between computational efficiency and segmentation accuracy, all CBCT images are normalized to $0.4 \times 0.4 \times 0.4 mm^3$. Then, in order to reduce the influence of extreme values, especially in the metal artifact area, we clip the voxel wise intensities value of each CBCT scan to [0, 2500], and finally normalize the pixel value to the interval [0, 1]. Similar to previous studies \cite{8954147,10.1007/978-3-030-78191-0_12,article}, we used typical segmentation metrics for performance evaluation: Dice, Jaccard similarity coefficient (Jaccard), 95\% Hausdorff distance (HD95), Average surface distance (ASD) and Sensitivity.

\subsection{Performance Comparison}
\begin{table}
\caption{Performance comparison of different methods}
\setlength{\tabcolsep}{1pt}
\resizebox{\columnwidth}{!}{
\begin{tabular}{c|c|ccccc}
\hline
Methods & Class & Dice($\%$)$\uparrow$ & Jaccard($\%$)$\uparrow$ & HD95$\downarrow$ & ASD$\downarrow$ & Sensitivity($\%$)$\uparrow$\\
\hline
ToothNet\cite{8954147} & ~ & 91.45$\pm$1.06 & 84.21$\pm$1.71	& 2.96$\pm$1.43 &	0.31$\pm$0.13 & 89.60$\pm$1.52\\
CGDNet\cite{9098542} & ~ & 92.31$\pm$0.57 & 85.11$\pm$1.53 & 2.30$\pm$1.05 & 0.29$\pm$0.09 & 90.34$\pm$1.93\\
ToothSeg\cite{10.1007/978-3-030-78191-0_12} & Tooth & 93.45$\pm$0.46 & 88.92$\pm$0.92 & 1.68$\pm$0.21 & 0.25$\pm$0.05 & 91.38$\pm$1.32\\
Ours & ~ & 94.47$\pm$0.45 & 90.35$\pm$0.73 & 1.47$\pm$0.26 & 0.21$\pm$0.05 & 93.83$\pm$1.31\\
\hline
Methods & Class & Dice($\%$)$\uparrow$ & Jaccard($\%$)$\uparrow$ & HD95$\downarrow$ & ASD$\downarrow$ & Sensitivity($\%$)$\uparrow$\\
\hline
CGDNet\cite{9098542} & ~ & 69.71$\pm$3.08&	56.41$\pm$2.53	&3.55$\pm$1.71&	5.63$\pm$2.13	&88.75$\pm$1.46\\
ToothSeg\cite{10.1007/978-3-030-78191-0_12} & Root & 71.78$\pm$3.32	&58.31$\pm$2.44&	2.89$\pm$1.45	&4.94$\pm$2.02&	89.41$\pm$1.37\\
RCS\cite{9629727} & canal & 70.37$\pm$2.71&	56.89$\pm$2.81&	3.45$\pm$1.66&	5.71$\pm$1.95	&87.35$\pm$1.40\\
Ours & ~ & 72.92$\pm$2.73&	58.71$\pm$2.36&	2.21$\pm$1.31&	4.34$\pm$1.88	&92.11$\pm$1.21\\
\hline
\end{tabular}}
\label{tab1}
\end{table}

\begin{table}
\caption{Ablation study. WT represents weight transfer, ZXY represents ZXYformer, AB represents auxiliary branch for uncertainty estimation.}
\setlength{\tabcolsep}{1pt}
\resizebox{\columnwidth}{!}{
\begin{tabular}{c|ccc|ccccc}
\hline
Class & WT & ZXY& AB & Dice($\%$)$\uparrow$ & Jaccard ($\%$)$\uparrow$ & HD95$\downarrow$ & ASD$\downarrow$ & Sensitivity($\%$)$\uparrow$\\
~&~&~&~&		91.32$\pm$0.73	&83.04$\pm$1.05	&3.02$\pm$1.78	&0.32$\pm$0.13	&88.41$\pm$1.47\\
~&$\surd$&~&~&			91.87$\pm$0.45	&84.72$\pm$0.83&	2.71$\pm$0.58&	0.28$\pm$0.07&	90.54$\pm$1.51\\
Tooth&~&$\surd$&~&		93.51$\pm$0.51	&88.90$\pm$1.01&	1.68$\pm$0.34&	0.23$\pm$0.05&	91.41$\pm$1.29\\
~&~&~&$\surd$&	92.91$\pm$0.63	&87.91$\pm$0.97&	1.77$\pm$0.28&	0.27$\pm$0.05	&90.09$\pm$1.61\\
~&$\surd$&$\surd$&$\surd$&	94.47$\pm$0.45&	90.35$\pm$0.73	&1.47$\pm$0.26&	0.21$\pm$0.05	&93.83$\pm$1.31\\
\hline
~&~&~&~&		69.43$\pm$3.41	&56.37$\pm$2.97&	3.92$\pm$1.84	&5.90$\pm$2.34	&84.90$\pm$1.57\\
Root&$\surd$&~&~&			70.88$\pm$3.22	&57.18$\pm$2.45&	3.12$\pm$1.73&	5.02$\pm$2.44&	89.09$\pm$1.61\\
canal&~&$\surd$&~&		72.07$\pm$2.70	&58.21$\pm$2.43	&2.53$\pm$1.42&	4.70$\pm$2.28	&91.01$\pm$2.01\\
~&~&~&$\surd$&	71.30$\pm$2.95	&57.53$\pm$2.78	&2.87$\pm$1.51	&4.84$\pm$2.07	&86.27$\pm$1.45\\
~&$\surd$&$\surd$&$\surd$&	72.92$\pm$2.73	&58.71$\pm$2.36	&2.21$\pm$1.31	&4.34$\pm$1.88	&92.11$\pm$1.21\\
\hline
\end{tabular}}
\label{tab2}
\end{table}

Table \ref{tab1} shows the performance comparison between the proposed method and other methods in teeth segmentation and root canals segmentation. Ref \cite{8954147} based on edge map supervision and Ref \cite{9098542} based on tooth center can achieve good performance, while Ref \cite{10.1007/978-3-030-78191-0_12} based on tooth shape can achieve better performance. By comparison, the proposed method is based on tooth and root canal morphology, as well as local and global attention of channel, space and depth, yielding the best performance. In addition, Table \ref{tab2} shows the ablation study performance of the proposed method. It can be observed from Table \ref{tab2} that the performance of teeth segmentation can be improved by transferring the weight of macro information in the coarse segmentation network to the fine segmentation network. Then, the experimental results also verify the effectiveness of the designed ZXYformer. In addition, for difficult areas such as tooth boundary and root canal calcification between teeth and alveolar bone, the proposed uncertainty module can well capture the areas to be refined, and improve the segmentation performance of the model for difficult areas by minimizing the uncertainty of prediction.

\begin{figure}[htb]
\centerline{\includegraphics[width=\columnwidth]{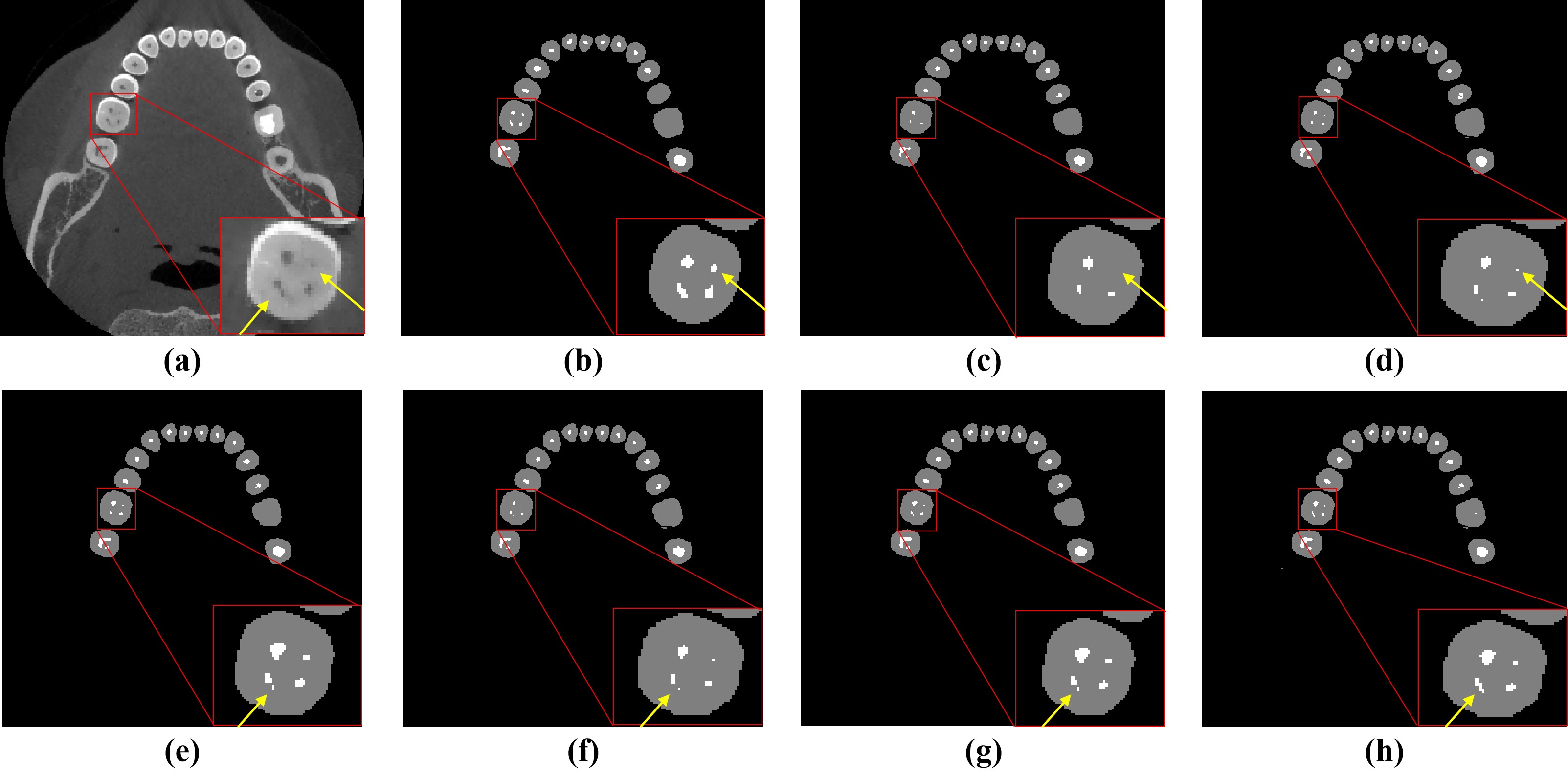}}
\caption{Visualization of results. (a) Original drawing, (b)Grond truth, (c) Ref \cite{9098542}, (d) Ref \cite{9629727}, (e) Method \cite{10.1007/978-3-030-78191-0_12}, (f) Weight Transfer is used in the proposed model (g) Weight transfer and ZXYformer is used in the proposed model (h) The proposed whole model.}
\label{fig3}
\end{figure}

From the upper right root canal of the enlarged picture in Fig \ref{fig3}(c), it can be seen that the tiny root canal is difficult to segment only based on the tooth center guidance, and the performance can be slightly improved after the root canal boundary supervision (d) and the tooth morphology guidance (e). The proposed method can obtain better segmentation results after introducing macro information and ZXYformer. In addition, the left lower root canal is partially calcified, and the segmentation results of other methods are discontinuous. The proposed method clearly improves the segmentation performance of this difficult areas after introducing uncertainty supervision.

\begin{figure}[htb]
\centerline{\includegraphics[width=\columnwidth]{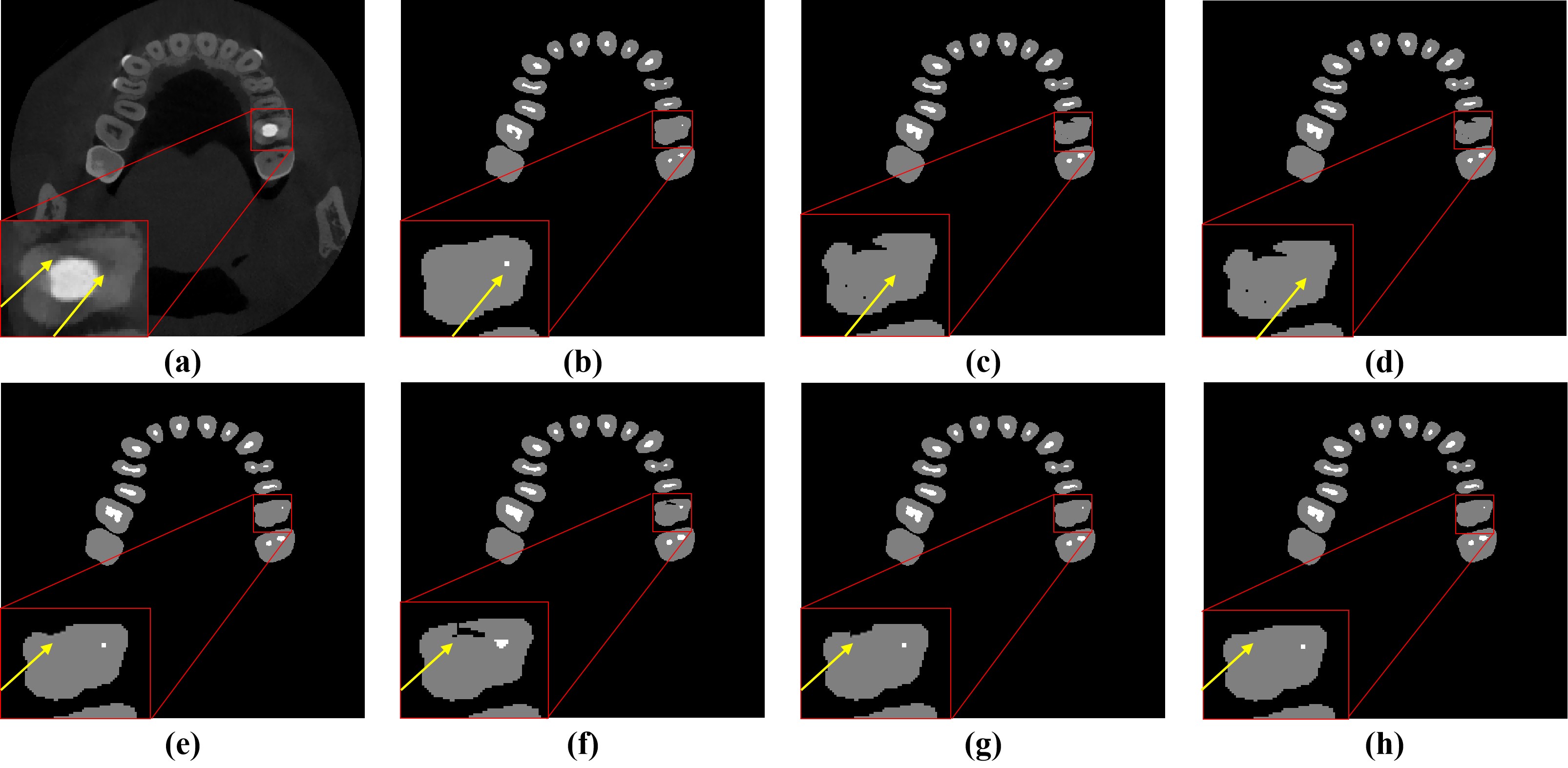}}
\caption{Visualization of results. (a) Original drawing, (b)Grond truth, (c) Ref \cite{8954147}, (d) Ref \cite{9098542}, (e) Ref \cite{10.1007/978-3-030-78191-0_12}, (f) Weight Transfer is used in the proposed model (g) Weight transfer and ZXYformer is used in the proposed model (h) The proposed whole model.}
\label{fig4}
\end{figure}

Fig. \ref{fig4} (a) is an enlarged view of a tooth that has undergone root canal treatment, in which the high-signal part is the implanted dental restoration, and the original root canal (white dot in Fig. \ref{fig4}(b)) still exists, and this tooth is accompanied by serious tooth decay (low-signal part in the tooth). As shown in Fig.4(d), Ref \cite{9098542} cannot make a correct judgment on tooth decay and only a few root canal divisions based on the tooth center or edge. Ref \cite{10.1007/978-3-030-78191-0_12} after adding morphological information, the segmentation of root canal and partial cavities has been improved to some extent, as shown in Fig. \ref{fig4}(e). After introducing weight transfer and ZXYformer, the proposed method learned more macro information and morphological information of teeth and root canals, and improved the tiny root canals and cavities. Finally, the difficult areas of cavities was optimized by dividing the uncertainty to achieve the best results, as shown in Fig. \ref{fig4}(h).

\section{Conclusion}
In this work, we proposed a ZXYformer with uncertainly guidance and macro weight transfer to improve the performance of joint teeth and root canals segmentation, outperforming previously reported teeth and root canals segmentation methods. We believe that the proposed method will be helpful to the application of multi-target segmentation of high-resolution images in other fields.

\bibliographystyle{IEEEbib}
\bibliography{strings,refs}

\begin{thebibliography}{10}

\bibitem{0dd57b9b00f14995a680f4ccd419106f}
{Frank C.} Setzer, {Katherine J.} Shi, Zhiyang Zhang, Hao Yan, Hyunsoo Yoon,
  Mel Mupparapu, and Jing Li,
\newblock ``Artificial intelligence for the computer-aided detection of
  periapical lesions in cone-beam computed tomographic images,''
\newblock {\em Journal of Endodontics}, vol. 46, no. 7, pp. 987--993, July
  2020,
\newblock Publisher Copyright: {\textcopyright} 2020 American Association of
  Endodontists.

\bibitem{8954147}
Zhiming Cui, Changjian Li, and Wenping Wang,
\newblock ``Toothnet: Automatic tooth instance segmentation and identification
  from cone beam ct images,''
\newblock in {\em 2019 IEEE/CVF Conference on Computer Vision and Pattern
  Recognition (CVPR)}, 2019, pp. 6361--6370.

\bibitem{10.1007/978-3-030-78191-0_12}
Zhiming Cui, Bojun Zhang, Chunfeng Lian, Changjian Li, Lei Yang, Wenping Wang,
  Min Zhu, and Dinggang Shen,
\newblock ``Hierarchical morphology-guided tooth instance segmentation from
  cbct images,''
\newblock in {\em Information Processing in Medical Imaging}, Aasa Feragen,
  Stefan Sommer, Julia Schnabel, and Mads Nielsen, Eds., Cham, 2021, pp.
  150--162, Springer International Publishing.

\bibitem{article}
Zhiming Cui, Yu~Fang, Lanzhuju Mei, Bojun Zhang, Bo~Yu, Jiameng Liu, Caiwen
  Jiang, Yuhang Sun, Lei Ma, Huang Jiawei, Yang Liu, Yue Zhao, Chunfeng Lian,
  Zhongxiang Ding, and Min Zhu,
\newblock ``A fully automatic ai system for tooth and alveolar bone
  segmentation from cone-beam ct images,''
\newblock {\em Nature Communications}, vol. 13, pp. 2096, 04 2022.

\bibitem{9098542}
Xiyi Wu, Huai Chen, Yijie Huang, Huayan Guo, Tiantian Qiu, and Lisheng Wang,
\newblock ``Center-sensitive and boundary-aware tooth instance segmentation and
  classification from cone-beam ct,''
\newblock in {\em 2020 IEEE 17th International Symposium on Biomedical Imaging
  (ISBI)}, 2020, pp. 939--942.

\bibitem{9629727}
Jian Zhang, Wenjun Xia, Jiaqi Dong, Zisheng Tang, and Qunfei Zhao,
\newblock ``Root canal segmentation in cbct images by 3d u-net with global and
  local combination loss,''
\newblock in {\em 2021 43rd Annual International Conference of the IEEE
  Engineering in Medicine and Biology Society (EMBC)}, 2021, pp. 3097--3100.

\bibitem{9759447}
Lei Ding, Dong Lin, Shaofu Lin, Jing Zhang, Xiaojie Cui, Yuebin Wang, Hao Tang,
  and Lorenzo Bruzzone,
\newblock ``Looking outside the window: Wide-context transformer for the
  semantic segmentation of high-resolution remote sensing images,''
\newblock {\em IEEE Transactions on Geoscience and Remote Sensing}, vol. 60,
  pp. 1--13, 2022.

\bibitem{XIA2022102581}
Likun Xia, Hao Zhang, Yufei Wu, Ran Song, Yuhui Ma, Lei Mou, Jiang Liu, Yixuan
  Xie, Ming Ma, and Yitian Zhao,
\newblock ``3d vessel-like structure segmentation in medical images by an
  edge-reinforced network,''
\newblock {\em Medical Image Analysis}, vol. 82, pp. 102581, 2022.

\bibitem{8237351}
Jifeng Dai, Haozhi Qi, Yuwen Xiong, Yi~Li, Guodong Zhang, Han Hu, and Yichen
  Wei,
\newblock ``Deformable convolutional networks,''
\newblock in {\em 2017 IEEE International Conference on Computer Vision
  (ICCV)}, 2017, pp. 764--773.

\bibitem{10.1007/s11263-020-01395-y}
Zhedong Zheng and Yi~Yang,
\newblock ``Rectifying pseudo label learning via uncertainty estimation for
  domain adaptive semantic segmentation,''
\newblock {\em Int. J. Comput. Vision}, vol. 129, no. 4, pp. 1106–1120, apr
  2021.

\end{thebibliography}

\end{document}